\documentclass{ieeeaccess}
\usepackage{graphicx}
\usepackage{amsmath}
\usepackage{booktabs}
\usepackage{multirow}
\usepackage[cjk]{kotex}
\usepackage{xcolor}

\usepackage{amssymb}
\usepackage{pifont}
\usepackage{makecell}
\definecolor{cvprblue}{rgb}{0.21,0.49,0.74}
\usepackage[pagebackref,
            breaklinks,
            urlcolor=cvprblue,
            colorlinks=cvprblue,
            citecolor=cvprblue]{hyperref}
\usepackage[capitalize]{cleveref} 
\usepackage{cite}
\usepackage{amsmath,amssymb,amsfonts}
\usepackage{algorithmic}
\usepackage{textcomp}
\usepackage{subcaption}
\usepackage{lipsum}
\usepackage{colortbl}

\newcommand{\cmark}{\ding{51}}%
\newcommand{\xmark}{\ding{55}}%

\definecolor{lightgray}{gray}{0.9}

\newcommand{\mc}[2]{\multicolumn{#1}{c}{#2}}
\definecolor{Gray}{gray}{0.85}
\newcolumntype{a}{>{\columncolor{Gray}}c}

\newcommand{\orcid}[1]{\href{https://orcid.org/#1}{\includegraphics[width=8pt]{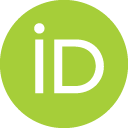}}}

\def\BibTeX{{\rm B\kern-.05em{\sc i\kern-.025em b}\kern-.08em
    T\kern-.1667em\lower.7ex\hbox{E}\kern-.125emX}}
\begin{document}
\history{Date of publication xxxx 00, 0000, date of current version xxxx 00, 0000.}
\doi{10.1109/ACCESS.2017.DOI}

\title{Exploring Syn-to-Real Domain Adaptation for Military Target Detection}
\author{
    \uppercase{Jongoh Jeong}$^{\orcid{0000-0002-5354-2693}}$\authorrefmark{1} ,\IEEEmembership{Graduate Student Member, IEEE}, \uppercase{Youngjin Oh}\authorrefmark{1}, \\
    \uppercase{Gyeongrae Nam}$^{\orcid{0009-0009-0873-2701}}$\authorrefmark{2}, \uppercase{Jeongeun Lee}$^{\orcid{0000-0002-1478-442X}}$\authorrefmark{2},
    \uppercase{and}
    \uppercase{Kuk-Jin Yoon}$^{{\orcid{0000-0002-1634-2756}}}$\authorrefmark{1}, \IEEEmembership{Member, IEEE}
}
\address[1]{Department of Mechanical Engineering, Korea Advanced Institute of Science and Technology (KAIST) (e-mail: \{jeong2, an970529, kjyoon\}@kaist.ac.kr)}
\address[2]{Drone Research and Development Laboratory, LIG Nex1, Seongnam-si 13486, South Korea (e-mail: \{krnam81, jeongeun.lee\}@lignex1.com)}
\tfootnote{
}
\tfootnote{This research is performed based on the cooperation with KAIST-LIG Nex1 Cooperation (Grant No.Y22-C0010).}


\corresp{Corresponding author: Kuk-Jin Yoon (e-mail: kjyoon@kaist.ac.kr).}

\begin{abstract}
    Object detection is one of the key target tasks of interest in the context of civil and military applications. In particular, the real-world deployment of target detection methods is pivotal in the decision-making process during military command and reconnaissance. However, current domain adaptive object detection algorithms consider adapting one domain to another similar one only within the scope of natural or autonomous driving scenes. Since military domains often deal with a mixed variety of environments, detecting objects from multiple varying target domains poses a greater challenge. 
    Several studies for armored military target detection have made use of synthetic aperture radar (SAR) data due to its robustness to all weather, long range, and high-resolution characteristics. Nevertheless, the costs of SAR data acquisition and processing are still much higher than those of the conventional RGB camera, which is a more affordable alternative with significantly lower data processing time. Furthermore, the lack of military target detection datasets limits the use of such a low-cost approach. To mitigate these issues, we propose to generate RGB-based synthetic data using a photorealistic visual tool, Unreal Engine, for military target detection in a cross-domain setting. To this end, we conducted synthetic-to-real transfer experiments by training our synthetic dataset and validating on our web-collected real military target datasets. We benchmark the state-of-the-art domain adaptation methods distinguished by the degree of supervision on our proposed train-val dataset pair, and find that current methods using minimal hints on the image (e.g., object class) achieve a substantial improvement over unsupervised or semi-supervised DA methods. From these observations, we recognize the current challenges that remain to be overcome. Dataset will be released to public upon acceptance at \textsc{\href{}{LINK}}.
\end{abstract}

\titlepgskip=-15pt

\maketitle

\section{Introduction}
\label{sec:introduction}

\begin{figure*}[!t]
    \centering
    \includegraphics[width=.92\linewidth]{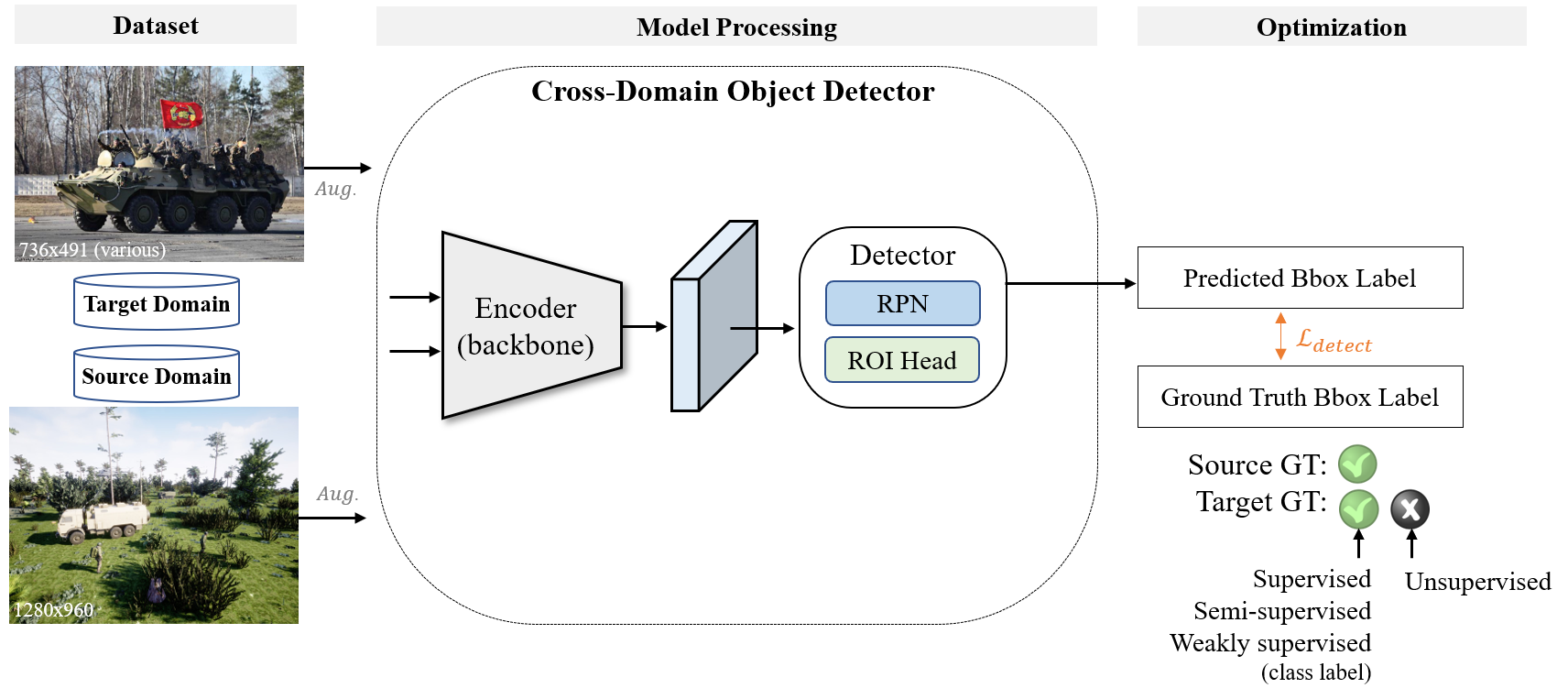}
    \caption{\textbf{Illustration of general domain adaptive object detection approaches}. 
    we explore supervised, semi-supervised, weakly supervised and unsupervised domain adaptation approaches for military target detection.}
    \label{fig:overview}
\end{figure*}

Deep learning techniques have drawn significant attention and have found numerous applications in civil and military applications, particularly for detection and tracking of objects in the wild~\cite{zheng2020mcships, meng2022visual, cazzato2020survey, vs2022meta, lo2021dynamic, zhang2022recognition, walambe2021multiscale}, as search and reconnaissance can be significantly enhanced with visual assistance. In the identification and localization of armored objects in the real world, objects to be found tend to be diverse in color, size, texture, and position, and the environments in which military arms are deployed vary in climatic characteristics such as weather, terrain, and whether there are surrounding objects for cover and concealment. Moreover, data processing in real time is indispensable for military applications, especially due to limited hardware resources in deployment, which already involve numerous parameters and calculations~\cite{calderon2020visual, mandal2019sssdet, wang2020ship}. 

A number of vision-based systems~\cite{wang2019change, yue2020novel, pannu2020deep, kechagias2021automatic, nemni2020fully, zhang2021multitask} have therefore presented feasible solutions to meet these needs by incorporating processed synthetic aperture radar (SAR) data as a tool for robust detection of target objects with lower resource costs. By exploiting millimeter wave radar scattering shot from the satellites, SAR data are advantageous on several measures: precise and high-resolution imagery, reliable and anywhere-on-earth imaging, and accurate 3D elevation modeling. However, SAR data are also constrained by limited access to satellite data, rigorous data processing, and the synthesis of multiple radar scans. Furthermore, the SAR modality is known for the lack of training data and the difficulty in ensuring its generalization capacity~\cite{zhao2000synthetic}, which is the key to adapting to diverse domains in the real world.

In response to these constraints, LiDAR and RGB data modalities alleviate the data processing burden. In particular, for military applications in which orders and actions must be put into action swiftly, the RGB modality serves as an adequate data format for armored object detection in such cross-domain setup. With the growing interest in domain adaptation (DA) algorithms using the RGB camera, image-based armored target detection task sets the ground for a practical potential military application scenarios.

In this regard, we propose to craft an RGB image-based synthetic dataset as well as a paired real image dataset for armored target detection, following the general detection approach as illustrated in Fig.~\ref{fig:overview}. We build our synthetic dataset on a physics engine-based simulator for photo-realistic visuals suited to military settings (e.g., wild forest, desert landscape). The captured images from the simulation contain assets belonging to the representative military target classes, which we defined according to the availability of real dataset on the Internet. We ensure that the label categories for the synthetic and real datasets match, yet the domain distributions are dissimilar. While the real images inherently pose diverse appearances due to the nature of collection process on the Web, they are also well spread within each class, e.g., a tank shot from a height above at a downward angle, or a low-resolution close-up look of a tank. Based on these characteristics, we contrive a challenging synthetic-to-real image dataset pair for armored military target detection and validate its applicability to image-based domain adaptive frameworks. 

\noindent The main contributions of this paper are as follows:
\begin{enumerate}
    \item To the best of our knowledge, we introduce a publicly available RGB-based military target detection dataset in synthetic environments to the research community for the first time. Paired with the proposed synthetic dataset, we also provide a collection of web-collected real image data for evaluation.
    \item We provide benchmark experimental results for comparison across various domain adaptation strategies to give insights on the difficulties and limitations that still remain as challenges in cross-domain object detection.

    \item We find that weak supervision (class labels) significantly helps boost the armored object detection accuracy compared to the unsupervised approaches under our challenging cross-domain setting.
\end{enumerate}

The rest of our paper is organized as follows: we introduce datasets and strategies for armored target detection in Sec.~\ref{sec:relatedwork}, elaborate on our dataset construction in Sec.~\ref{sec:dataset}, analyze experimental results in Sec.~\ref{sec:experiment}, and, lastly, conclude with remarks and future directions in Sec.~\ref{sec:conclusion}.

\section{Related Work}
\label{sec:relatedwork}

In this section, we summarize concurrent works on military object datasets and object detection methods in the context of domain adaptation in detail.

\begin{table*}[!t]
    \setlength{\tabcolsep}{1.5pt}
    \centering
    \caption{\textbf{Comparison of datasets for armored target detection.}}
    \label{tab:dataset_comparison}
    \resizebox{\linewidth}{!}{
        \begin{tabular}{lccccccccccc}
        \toprule
             & & & & \multicolumn{3}{c}{\textbf{Modality}} \\ \cmidrule{5-7}
             \textbf{Dataset} & \textbf{Year} & \textbf{Data Source} & \textbf{Available?} & RGB & LiDAR & SAR & \textbf{Anno. Type} & \textbf{\# Anno. Instances} & \textbf{\# Samples} & \textbf{\# Classes} & \textbf{Resolution} \\
        \midrule 
             MSTAR~\cite{DARPA1996mstar} & 1996 & Real (satellite) & \cmark && & \cmark & Class label & 20,000 & 20,000 & 10 & 0.3 m (Rg.)$\times$ 0.3 m (Az.) \\
             OpenSARShip~\cite{openSARship,openSARship2} & 2017 & Real (satellite) & \cmark & & & \cmark & Bounding box & 11,346 ships & 11,346 & 17 & 1.7 m (Rg.) ~ 22 m (Az.) \\
             \cite{xiaozhu2017object} & 2017 & Synthetic (web) & \xmark & \cmark & & & Bounding box & \textcolor{gray}{N/A} & 6,000 & 2 & 500$\times$375 (pix)\\
             MGTD~\cite{Belloni2019MGTD_SAR} & 2019 & Real (satellite) & \xmark & & & \cmark & Bounding box & \textcolor{gray}{N/A} & 1,728 & \textcolor{gray}{N/A} & 0.03 m (Rg.) $\times$0.033 m (Az.) \\ 
             ARTD~\cite{MENG2020SAR_ARTD} & 2020 & Synthetic (video game, web) & \xmark & \cmark && & Bounding box & 30,132 & 11,536 & 30,132 & 10$\times$10 $\sim$ 700$\times$700 (pix)\\
             \cite{dai2022gcd}, \cite{zhao2023lidar} & 2022 & Real & \xmark & & \cmark & & Bounding box & \textcolor{gray}{N/A} & 3,000 & 3 & \textcolor{gray}{N/A} \\ 
             \cite{du2022lightweight} & 2022 & Synthetic (web) & \cmark (upon request only) & \cmark & & & Bounding box & 13,199 & 9,369 & 7 & [167$\times$2802] $\sim$ [233$\times$4960] (pix) \\
             \midrule
             
             \rowcolor{lightgray}
             \textbf{Ours} & \textbf{2023} & \textbf{Synthetic (simulated) / Real (web)} & \cmark & \cmark && & \textbf{Bounding box} & \textbf{14,944 / 1,781} & \textbf{5,996 / 774} & \textbf{6} & \textbf{1280$\times$680 / 800$\times$640} (pix) \\
        \bottomrule
        \end{tabular}
    } 
\end{table*}

\subsection{Armored Military Target Datasets}
    In this subsection, we detail existing datasets designed for military target recognition, i.e., detecting objects in bounding boxes.

    \subsubsection{RADAR data} \quad
    Datasets for military target detection often involve synthetic aperture radar (SAR) imagery data to be robust to varying environmental conditions around the target objects. In contrast to conventional RGB cameras, SAR deals with collecting data from a satellite. That is, the data is essentially the amount of energy reflected back from the target on the Earth, thus involving response to surface characteristics such as terrain structure and moisture conditions. As opposed to common optical sensing, acquiring radar data of a sufficient spatial resolution would require an excessively long radar antenna (e.g., for a C-band radar with a wavelength of 5 centimeters, a spatial resolution of 10 meters requires a radar antenna of 4,250 meters long). As a remedy, a sequence of signals from shorter antennas are combined to provide higher-resolution data. SAR data acquisition often involves tedious data processing stages including applying orbit file, radiometric calibration, debursting, multi-looking, speckle filtering, and terrain correction. Moreover, a specialized software tool is necessary to process and visualize data. After data processing, a final ``SAR image", i.e., a grayscale or multiband image of radiometric terrain correction (RTC)-calibrated SAR backscatter, is produced~\cite{Belloni2019MGTD_SAR, SAR2019fundamental, gu2021SAR, zhou2021SAR, jiang2023SAR, Amrani2021SAR, belloni2017SAR, geng2023SAR, schumacherSAR2005, openSARship, openSARship2}. 
    We compare the dataset specifications in detail in Table~\ref{tab:dataset_comparison}.

    \noindent\textbf{MSTAR Dataset.} \quad
    Moving and Stationary Target Acquisition and Recognition (MSTAR) dataset~\cite{DARPA1996mstar} is a standard public-release dataset designed by a joint Defense Advanced Research Projects Agency (DARPA) and Air Force Research Laboratory (AFRL) primarily for an advanced automatic target recognition (ATR) system. It consists of nearly 20,000 airborne SAR images of ten various former Soviet Union vehicle types in 128$\times$ 128 resolution, observed in pose estimator ranging between 0$^{\circ}$ to 360$^{\circ}$, full 360$^{\circ}$ azimuth angles with 1$^{\circ}$ to 5$^{\circ}$ increments over target elevation (depression) angles of 15$^{\circ}$, 17$^{\circ}$, 30$^{\circ}$ and 45$^{\circ}$ , where the majority contains mostly 15$^{\circ}$ (3,423 images) and 17$^{\circ}$ (3,451 images). Due to the synthetic nature, radar data on each vehicle from a number of different longitudinal angles are combined to produce one whole image. Underlying assumptions include flat terrain surface and no occlusions on the vehicle. Each image chip sample is then paired with a unique label from a finite set of labels. The SAR images in the database are derived from X-band SAR images with the resolution of 1 ft $\times$ 1 ft in spotlight mode. A similar Military Ground Targets Dataset (MGTD)~\cite{Belloni2019MGTD_SAR} is introduced to address the bias present in MSTAR.
    
    As a standard approach adopted by many previous literature, the dataset is used for a 10-way vehicle classification problem with 17-degree longitudinal angle data as training images and 15-degree as test data~\cite{morgan2015SARimagery}.  Differently from the standard, \cite{zhao2000synthetic} uses 80$\times$80 SAR images from the MSTAR database on three specific ground targets, with 12 sector classifiers, each divided into 30-degree sector. \cite{sun2007adaptive} additionally assumes the MSTAR data are rectangular in shape; that is, the pose of the target vehicle is distinctly determined by the longer edge, where the target pose is defined as the angle between the longer side of the vehicle and the horizontal axis.

    \noindent\textbf{OpenSARShip Database.} \quad
    OpenSARShip~\cite{openSARship} is another radar-based maritime target detection dataset that exploits the availability of 41 different Sentinel-1 SAR imagery data sets collected from five scenes located at Shanghai Port (China), Shenzhen Port (China), Tianjin Port (China), Yokohama Port (Japan), and Singapore Port (Singapore), consisting of 11,346 ship chips integrated into 17 different automatic identification system (AIS) messages. Its original data are derived from the Sentinels Scientific Data Hub~\cite{ESA2015SAR} and the processed data are publicly available at the OpenSAR platform~\cite{shanghai2017openSARplatform}. The proposed OpenSARShip is said to exhibit five characteristics, namely specificality, large scale, diversity, reliability, and public availability. As the data undergoes four rigorous processing (raw data collection and pre-preocessing — online-coordination labeling tool — AIS message integration — post-processing to guarantee reliability and quality), data acquisition requires massive computational resources. 
    Extending this large-scale SAR imagery interpretation dataset, OpenSARShip 2.0~\cite{openSARship2} presents a deeper SAR imagery interpretation into the maritime target detection. It contains 34,528 SAR ship chips with AIS information and is improved in three aspects from the previous version: large volume, interference labeling, and type levels. Although these datasets are not originally designed for military target detection, their common motive and objective are well aligned with that of an armored target detection dataset. 
    
    \subsubsection{LiDAR Data}
        \noindent\cite{dai2022gcd, zhao2023lidar} present another armored target detection dataset, in which the simulated LiDAR and infrared (IR) data on armored targets at 60--100 meters in altitude are obtained in different terrains and angles. The three target classes include \textit{armored targets}, \textit{obstacles}, and \textit{ground background}, with 3,000 samples in total. In data acquisition, the LiDAR and IR sensors are attached to the main computer, fixed in position, and the objects of interest are shot from these sensors out in the wild.
        
    \subsubsection{RGB Data}
        \noindent RGB camera has the benefit of being cheaper and easier to access than RADAR or LiDAR sensors. Moreover, the sensor output is more intuitive to interpret and process than the other sensor outputs. Due to these advantages, \cite{xiaozhu2017object} collects 6,000 battlefield scene images of moving armored vehicles, specifically \textit{tank} and \textit{wheeled infantry fighting vehicle}, along with Pascal VOC format bounding box annotations. The RGB images are scaled to 500$\times$375, and each image contains real battlefield environments with various background items in various ranges, angles, and scales. With Faster R-CNN~\cite{ren2015fasterrcnn}, the detection rate reaches a fairly high accuracy given the RGB input only. Similarly, \cite{hao2017image} crafts a collection of armored tank target detection images to put to use with a hierarchical multi-scale detector model. 
        
        \noindent\textbf{Armored Target Dataset (ARTD).} \quad
        \cite{MENG2020SAR_ARTD} also introduced a dedicated RGB dataset for armored target detection which contains armored targets of particular types and complex battlefield scenes. It consists of 11,536 images and 30,132 unique armored targets in battlefield scenes (e.g., jungle, desert, grasslands, and city) taken from video games and the Web. It is uniquely designed to include complicated factors such as armored cluster, muzzle fire, and smoke to fill the data gap with concurrent datasets that contain a limited number of target classes and fewer samples.

\subsection{Domain Adaptive Object Detection}
    
    In this subsection, we briefly describe the methods we have experimented with in this paper, as illustrated in Fig.~\ref{fig:overview}. 
    
    Domain adaptation (DA) refers to learning a model trained on the source domain not to experience negative transfer on the target domain when the source and target domains differ. In order to reduce the negative transfer caused by a domain gap, or difference in their marginal data distributions, several adaptation approaches address the cross-domain problem. These approaches are often branched and classified by how much and whether labeled target annotations are available. 
    
    Conventional supervised neural network training assumes the source and target domains to be derived from the same data distributions, e.g., the same camera intrinsic parameters. However, when the testing domain changes drastically from the source, the encoded data representations are often off-centered away from the well-trained feature space, thereby inducing negative transfer and forgetting of previously learned knowledge. Thus, the objective of domain adaptation is to find the optimal space for target domain feature alignment such that the learned source domain knowledge is preserved and becomes useful in the other domain. Moreover, label space matching is another key component for adaptation. In other words, alignment of the feature-label pair $\mathcal{L}=\{(x, y)\}$  across domains whose data are labeled differently in one domain than in the other is a crucial consideration for DA. In the following, we specifically consider closed set domain adaptation~\cite{farahani2021brief}, where the source and target domains follow the same labeling nomenclature, while the different data distributions still pose a domain gap.

    \noindent\textbf{Supervised DA.} \quad
    Supervised domain adaptation can be considered as a special branch of transfer learning~\cite{zhuang2020comprehensive, torrey2010transfer} with labels available in both the training and test domains. This procedure is framed in two stages: source-domain training and target-domain training with the trained source-task weight initialization, in order. For the object detection task, we can simply fine-tune a well-trained state-of-the-art model (e.g., YOLO~\cite{ultralytics2023yolov8}) on target data to achieve comparable or better performance than that in single-domain training. 

    As the state-of-the-art supervised domain adaptation methods, we employed EfficientDet~\cite{tan2020efficientdet}, Faster R-CNN~\cite{ren2015fasterrcnn}, and Yolo V8x~\cite{ultralytics2023yolov8} in our comparative study.
    EfficientDet is an
    
    \noindent\textbf{Semi-supervised DA.} \quad
    Semi-supervised approach is a variant of the aforementioned supervised method, with the difference lying in the percentage of supervision during training. The goal here is to use a finite number of labeled data $\{(x, y)\}$, as well as unlabeled data, $\mathcal{U}=\{x\}$. The idea of this mixed data strategy came about because the cost of accurately labeling all data is high. Since there is a limited number of target labels, pseudo-labels from a teacher, $\hat{x}$, can be used to assist in training~\cite{sohn2020simple}. For semi-supervised object detection, \cite{yu2019semi_supervised} additionally enforces consistency between labeled and unlabeled in addition to conventional supervised loss to overcome the lack of labeled data. \cite{liu2021unbiased} further integrates student-teacher mutual learning along with the focal loss and exponential moving average (EMA) to alleviate class and foreground-background imbalance problems.

    As the state-of-the-art semi-supervised domain adaptation methods, we used Unbiased Teacher~\cite{liu2021unbiased} in our comparative experiments.
    
    \noindent\textbf{Weakly supervised DA.} \quad
    Weakly supervised DA, or cross-domain weakly supervised object detection (CDWSOD), exploits coarse annotations instead of fine bounding box annotations for a holistic understanding of the input data in domain-shifting scenarios. This learning paradigm came about due to the high cost of human labeling in the target domain. Several approaches have adopted image-level feature alignment~\cite{inoue2018cross, hou2021informative} and instance-level cycle confusion~\cite{wang2021robust} to realize robust object detection accuracy under domain shift by time of day and cross-camera. Current state-of-the-art CDWSOD method~\cite{h2fa2022weakly} employs both image-level and instance-level alignment in a hierarchical manner to facilitate detection in different domains, using full supervision on the source and image-level (weak) supervision on the target domain.
    %
    For our experiments, we used H2FA R-CNN~\cite{h2fa2022weakly} for weakly supervised domain adaptation. 
    
    \noindent\textbf{Unsupervised DA.} \quad
    %
    Unsupervised DA is a task setting in which the source image-label pairs are available while only the target images are available. As a popular approach, previous works focused on distribution matching strategies using the maximum mean discrepancy (MMD)~\cite{gretton2012kernel} or a domain classifier network~\cite{ganin2016domain, long2015learning, long2016unsupervised, tzeng2017adversarial}. 
    Recently, pseudo-labelling~\cite{deng2021unbiasedmeanteacher, chen2022probabilisticteacher, li2022cross, cao2023cmt}-based Mean Teacher (MT)~\cite{tarvainen2017meanteacher} gained attention for their remarkable detection performance. These methods enforce the consistency over distilled pseudo-labels and student predictions to secure robustness, as revealed by \cite{deng2021unbiasedmeanteacher} that MT alone fails in cross-domain object detection. This self-training idea from semi-supervised learning is being actively exploited by researchers for the unsupervised domain adaptive object detection task.

    As the unsupervised domain adaptation methods, we employed the state-of-the-art Adaptive Teacher (AT)~\cite{li2022cross}, Probabilistic Teacher (PT)~\cite{chen2022probabilisticteacher}, and Contrastive Mean Teacher (CMT)~\cite{cao2023cmt} in our experiments.
    Unsupervise

\section{Dataset Construction Method}
\label{sec:dataset}

In this section, we describe the processes of asset preparation, data construction, and post-processing for our synthetic dataset in detail. Moreover, we explain the methodology for collecting real images belonging to our defined classes along with the corresponding bounding box labels.

\begin{table}[t!]
    \centering
    \caption{\textbf{Detailed instance-level breakdown} of our synthetically crafted  (``Synthetic") and web-collected (``Real") datasets. }
    \label{tab:dataset}
    \resizebox{\columnwidth}{!}{
    \begin{tabular}{rcccc}
    \toprule
         & \multicolumn{1}{c}{Synthetic (\texttt{train})} && \multicolumn{1}{c}{Real (\texttt{train})} & \multicolumn{1}{c}{Real (\texttt{val})} \\ 
     \cmidrule{2-2} \cmidrule{4-5}
         Resolution & 1280$\times$960 && \multicolumn{2}{c}{800$\times$640 (Median)} \\ 
     \cmidrule{2-2} \cmidrule{4-5}
         Category & \# Anno. Instances && \# Anno. Instances &  \# Anno. Instances \\
     \midrule
     \midrule
         Artillery & 695 && 108 & 28\\
         Drone &  2,379  && 113 & 38\\
         Helicopter &  3,473 && 151 & 48\\
         Soldier & 3,024 && 609 & 317\\
         Tank &  2,811 && 135 & 40\\
         Vehicle & 2,562  && 151 & 43\\
     \midrule
        \rowcolor{lightgray}
         Total & 14,944 && 1,267 & 514 \\
     \bottomrule
    \end{tabular}
    }
\end{table}

\begin{figure}[t!]
    \centering
    \begin{subfigure}[b]{\columnwidth}
       \includegraphics[width=\linewidth]{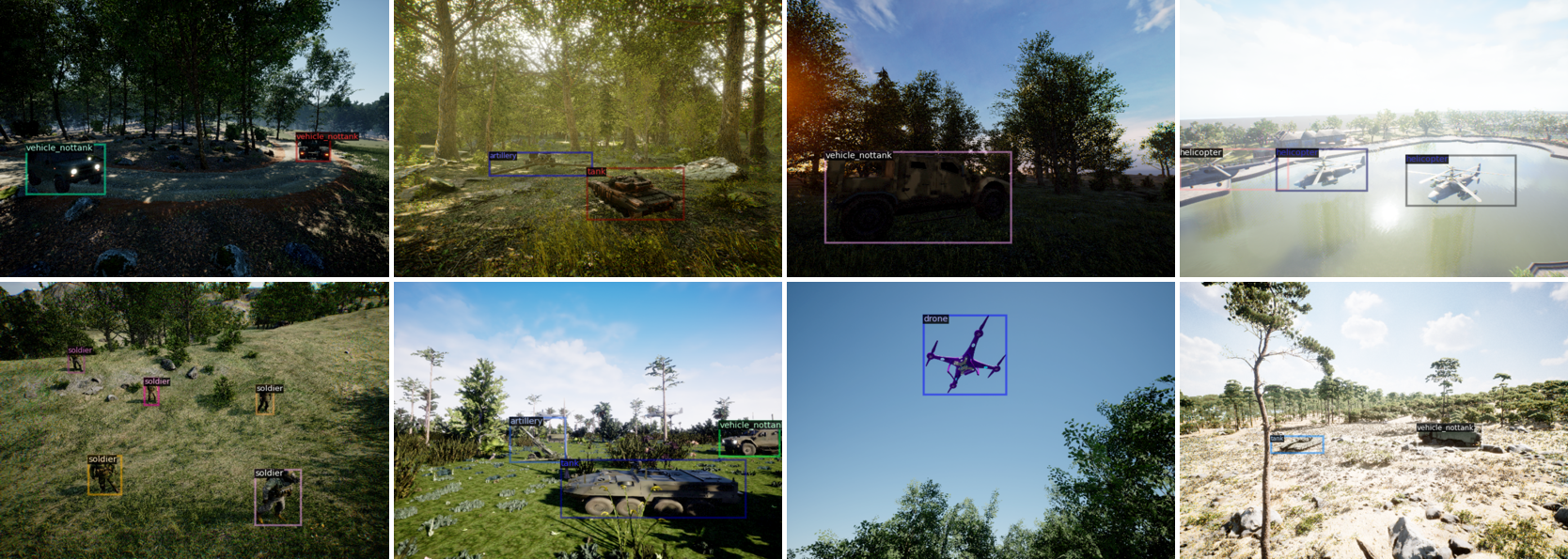}
       \caption{Synthetic}
       \label{fig:synthetic} 
    \end{subfigure}
    \begin{subfigure}[b]{\columnwidth}
       \includegraphics[width=\linewidth]{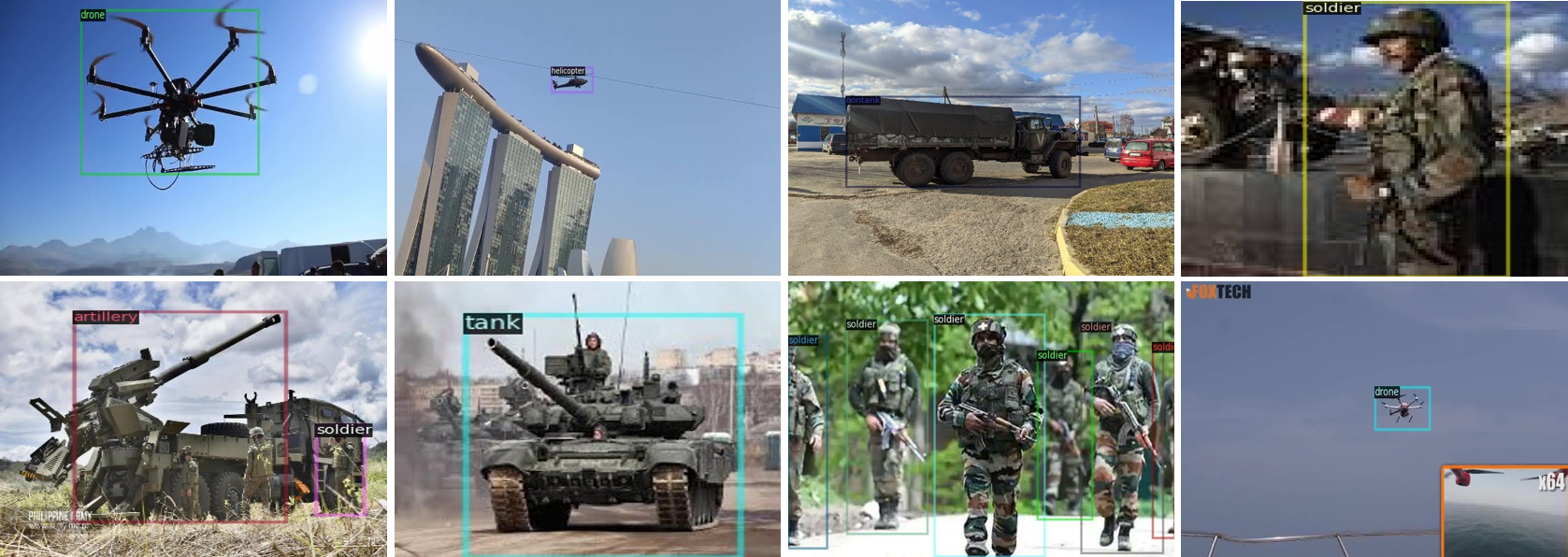}
       \caption{Real}
       \label{fig:real}
    \end{subfigure}
    \caption{\textbf{Selected image samples} from our synthetic (top) and real (bottom) datasets for demonstration. Bounding box annotations are accompanied in each dataset.}
    \label{fig:dataset_sample}
\end{figure}

\begin{figure*}[t!]
    \centering
    \includegraphics[width=.9\linewidth]{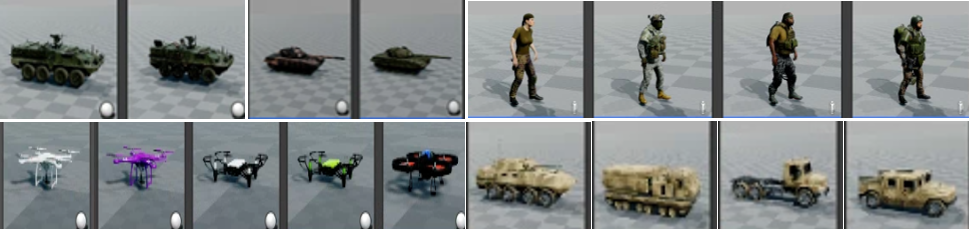}
    \caption{\textbf{Samples of the used simulator assets}: tank, armored soldiers, drones, and vehicles.}
    \label{fig:assets}
\end{figure*}

\begin{figure*}[!t]
    \centering 
    \begin{subfigure}{0.28\linewidth}
        \centering
      \includegraphics[width=\linewidth]{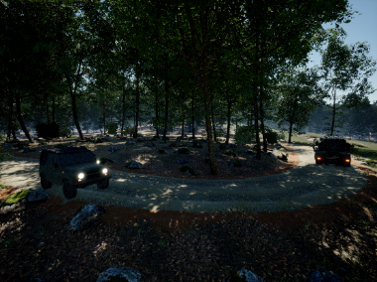}
      \caption{Oak Forest}
      \label{fig:1}
    \end{subfigure}\hfil 
    \begin{subfigure}{0.28\linewidth}
        \centering
      \includegraphics[width=\linewidth]{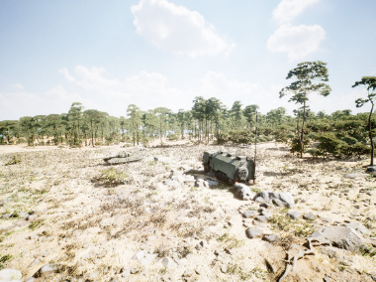}
      \caption{Mediterranean}
      \label{fig:2}
    \end{subfigure}\hfil 
    \begin{subfigure}{0.28\linewidth}
        \centering
      \includegraphics[width=\linewidth]{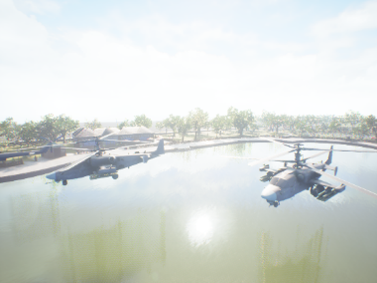}
      \caption{City Park}
      \label{fig:3}
    \end{subfigure}
    \medskip
    \begin{subfigure}{0.28\linewidth}
        \centering
      \includegraphics[width=\linewidth]{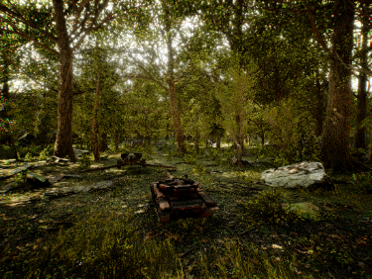}
      \caption{Woodland}
      \label{fig:4}
    \end{subfigure}\hfil 
    \begin{subfigure}{0.28\linewidth}
        \centering
      \includegraphics[width=\linewidth]{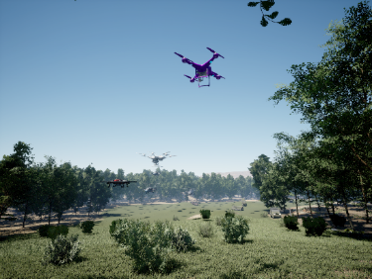}
      \caption{Forest}
      \label{fig:5}
    \end{subfigure}\hfil 
    \begin{subfigure}{0.28\linewidth}
        \centering
      \includegraphics[width=\linewidth]{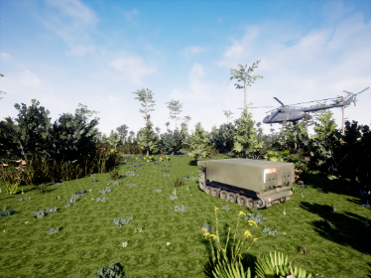}
      \caption{Baygall}
      \label{fig:6}
    \end{subfigure}
    \caption{\textbf{Samples of the simulated environments}: Oak Forest, Mediterranean, City Park, Woodland, Forest, and Baygall.}
    \label{fig:environments}
\end{figure*}

\begin{figure*}[!t]
    \centering 
    \includegraphics[width=\linewidth]{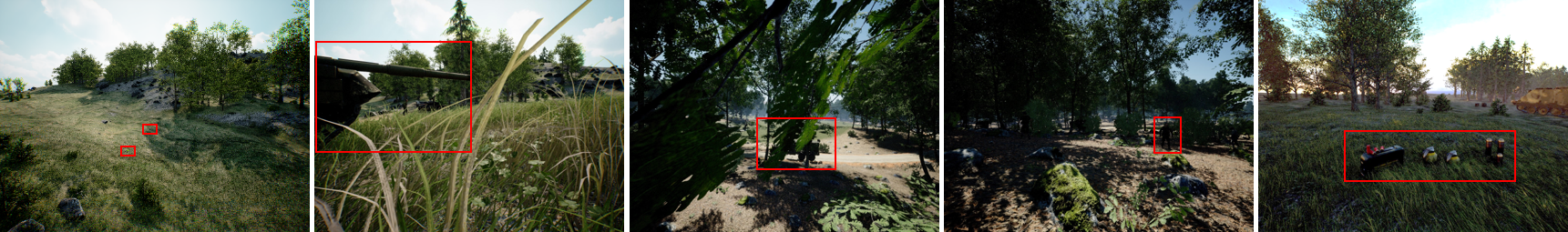}
    \caption{\textbf{Removed frames from the synthetic dataset}. Best viewed in \textit{zoom}.}
    \label{fig:removedframes}
\end{figure*}

    \subsection{Synthetic Dataset}
        \noindent\textbf{Class definition.} \quad
        We define our six military categories as \textit{drone, soldier, tank, vehicle, artillery} and \textit{helicopter} following a previous work~\cite{du2022lightweight}, which categorizes into seven unique classes. Although the classes defined in the previous work include maritime military vehicles (\textsc{warship}), we restrict our definition to ground military targets, as RGB cameras would typically be positioned at near-ground altitudes. 
        
        \noindent\textbf{Asset preparation.} \quad
        We run all asset and environment preparation and data generation in Unreal Engine 4.72.2~\cite{unrealengine}. We acquired the simulator objects, e.g., armed personnel, tank, helicopters, and drones, from the marketplace either for free or with the respective cost on the market. We provide the detailed asset names in the following: M777 Howitzer Artillery (West), Missile 9M55 (East), Missile Kh-29T (East), Missile AGM88 (West), Missile 9K121 Vihkr (East), Quadcopter pack - drones, UH60A Blackhawk (West), Mi8T Hip (East), Ka-52 Alligator (East), AH-64D Apache Longbow (West), MAE Oak Forest, City Park Environment Collection LITE, Mediterranean Island Pack, Brushify - Forest Pack (3), Baygall Brush, Military Character Pack, Simple Military, Tank T72B (East), M1133 Stryker MEV (West), SRMB Iskander(East), Military Armored Vehicle, M923 Truck (West), and Low Poly - Military Vehicles. Note that the some of the asset properties (\textit{i.e.,} outer texture, position, and scale) have been customized, and several assets may have been taken down from the market. References to the assets are available upon request.

        \noindent\textbf{Dataset construction.} \quad
        First, we constructed wild environments that could be encountered during military drills or missions. For instance, a group of armed infantry or a tank that pops out of the bushes for a guerilla attack would be found in a dense forest environment. In such armored situations, weaponry may also blend in with the surroundings. Therefore, we adjusted the texture and color to dissolve the target object in the background as cover and concealment, as seen in Fig.~\ref{fig:environments}. We also used Microsoft Airsim~\cite{shah2018airsim}, a plug-in tool that allows to navigate in either Drone, Car, or Computer Vision (first person's point of view) mode, and easily visualize depth and segmentation maps in real time. We used the Computer Vision mode to collect our dataset during manual navigation.
        
        As for the assets, we demonstrate select items in Fig.~\ref{fig:assets} that we included throughout our designed simulation settings. The employed assets include a variety of tanks, non-tank vehicles, armored soldiers, drones, and helicopters. The instances are positioned at appropriate locations under various illuminations, dense forests, desert-like environments. 
        The final labeled synthetic samples are visualized in Fig.~\ref{fig:dataset_sample}.

    \noindent\textbf{Post-processing.} \quad
    As deep learning networks are prone to over-fitting and sensitive to dataset quality and scale, our synthesized data must be verified of its quality as well as quantity. We eliminated samples according to the set of criteria defined below:
    \begin{enumerate}
        \item Extremely small size
        \item Stale and stagnant frames
        \item Unrecognizable object in whole (\textit{e.g.,} tank, soldier, vehicle)
        \item Does not belong in the defined class set (\textit{e.g.,} cartridge for ammunition, etc.)
    \end{enumerate}
    Following these guidelines, we show that several representative samples that have been removed in Fig.~\ref{fig:removedframes} and summarize the breakdown statistics at the instance level in Table~\ref{tab:dataset}. 


    \subsection{Real Dataset}        
        We acquired our collection of real data by compiling image-label pairs for each category from Roboflow as follows: Artillery\footnote{\scriptsize{\url{https://universe.roboflow.com/nbghzc/naveen}}}, Drone\footnote{\scriptsize{\url{https://universe.roboflow.com/icvf21iudronedetection/drones-qanmq}}}, Helicopter\footnote{\scriptsize{\url{https://universe.roboflow.com/illia-maliga-ojvda/helicopters-o0qig}}}, Soldier\footnote{\scriptsize{\url{https://universe.roboflow.com/military-civil-classification/edit_military}}}, Tank\footnote{\scriptsize{\url{https://universe.roboflow.com/geertspace/tanks-tsgjt}}}, Vehicle\footnote{\scriptsize{\url{https://universe.roboflow.com/itmo-3j5fx/mt-hzddy}}}.
        Since there exists no single dataset that contains ground objects related to our military setting, we compiled a separate dataset for each of the six classes, and then set the training set size to 600 and validation to 174 to be consistent across the classes. We acknowledge that a few samples in the real dataset may contain samples taken at undesirable angles or views. Nevertheless, they present challenging scenarios in which the sensor may have fallen off the fixed position or misplaced, thus helping us to evaluate on a few difficult scenes as well. We showcase the web-collected real samples in Fig.~\ref{fig:dataset_sample}, which we used for evaluation throughout our experiments.

\section{Experiments}
\label{sec:experiment}

\subsection{Model Overview}

    \noindent\textbf{Base object detector.} \quad
    We follow the standard cross-domain object detection training strategy by adopting the two-stage base object detector, Faster R-CNN~\cite{ren2015fasterrcnn}. As an improved extension of R-CNN~\cite{girshick2015region} and Fast R-CNN~\cite{girshick2015fast}, Faster R-CNN is a single-stage object detector, whose encoder backbone is widely used in various training frameworks, such as MaskRCNN benchmark~\cite{massa2018mrcnn}, Detecton2~\cite{wu2019detectron2}, FCOS~\cite{tian2019fcos, tian2021fcos}, due to its flexibility in design as a backbone. 
    %
    %
    %
    To elaborate, Faster-RCNN is a one-stage supervised object detector comprising three components: a backbone network, a region proposal network (RPN), and a region-of-interest (ROI)-based classifier. Using the extracted feature map from the backbone network, the RPN generates candidate object proposals using an anchor-based approach and constructs an ROI feature vector, $(x,y,w,h)$, of a fixed size using ROI pooling, where $(x,y)$ and $(w,h)$ each denote the upper left pixel coordinate and the ROI width and height. The ROI-based classifier then predicts the category and the bounding box for each proposed region. We can formulate the minimizing training objective to consist of the RPN loss and the ROI classifier loss as in the following equation:
    \begin{equation}
        \underset{}{\mathrm{argmin}}\;\mathcal{L}_{loc} + \mathcal{L}_{cls},
    \end{equation}
    \noindent where $\mathcal{L}_{loc}$ is a regression loss (\textit{i.e.,} $\mathcal{L}_{1}$ smooth loss) and $\mathcal{L}_{cls}$ is the classification loss (\textit{i.e.,} cross-entropy loss).
    

    \noindent\textbf{Benchmarked methods.}\quad 
    In the following paragraphs, we describe the method and our implementation details of the benchmark the current state-of-the-art object detection algorithms for other supervised, unsupervised, semi-supervised, weakly supervised and supervised domain adaptation.

    \noindent\textit{EfficientDet.}\quad   
    EfficientDet~\cite{tan2020efficientdet} is a convolutional neural network-based detection variant of EfficientNet~\cite{tan2019efficientnet} that uses a compound scaling approach to balance the model depth, width, and resolution via a compound coefficient. By this efficient scaling method that uniformly scales the three network hyper-parameters, the network manages a good trade-off between accuracy and computational efficiency. Extending this idea to object detection task, EfficientDet achieves a real-time performance by making use of a bi-directional feature pyramid network (BiFPN) and a compound scaling method to uniformly scale depth, width, and resolution for all backbones, feature networks and box/class detectors simultaneously.
    
    \noindent\textit{Yolo V8x.}\quad
    You Only Look Once (YOLO)~\cite{redmon2016yolo} is another acclaimed real-time network for one-stage object detector and image segmentation. Due to its remarkable speed and accuracy given its highly adaptable nature, the subsequent versions have come about, making it one of the most widely used algorithm in real-world applications. Version 8 (Yolo V8) improves upon previous anchor-free models on performance, flexibility, and efficiency measures, with a wide-range of supports for detection, image segmentation, pose estimation, tracking and classification for the ease of users. The benefit of extending anchor-free detection variants lies in predicting the center of an object directly instead of the offset from a known anchor box. To be more specific, Yolo v8, with its single forward pass operation, reduces computationally intensive operations of image scanning multiple times at different scales in traditional object detection methods. The adaptability of YOlo v8 to identify objects with high precision given images of tiny to large scale structures stands as a strong advantage to the traditional ones.

    

    \noindent\textit{Adaptive Teacher.} \quad
    Adaptive Teacher (AT)~\cite{li2022cross} is an unsupervised domain adaptation method that exploits the teacher-student framework to mitigate the problem of domain shift and avoid low-quality pseudo labels (\textit{i.e.,} false positives) in cross-domain settings. To that end, AT leverages feature-level domain adversarial learning for the student model, and weak-strong data augmentation and mutual learning between the teacher model and the student model to reduce domain gap without being biased to the source domain.

    
    \noindent\textit{Probabilistic Teacher.}\quad    
    Probabilistic Teacher (PT)~\cite{chen2022probabilisticteacher} follows AT in adopting the teacher-student learning framework and leveraging the uncertainty of pseudo boxes during self-training. The primary difference from the AT learning framework is to encourage mutual consistency between a gradually evolving teacher and a student guided by uncertainty of unlabeled target data, instead of filtering pseudo boxes thresholded by some confidence bound. 
    

    \noindent\textit{Contrastive Mean Teacher.}\quad   
    Contrastive Mean Teacher~\cite{cao2023cmt} finds the synergy between mean-teacher self-training and momentum contrast approaches by aligning them in a unified framework to learn maximally beneficial signals. Moreover, the addition of object-level contrastive learning on noisy pseudo-labels effectively trains the mean teacher network without requiring target domain labels.

    \noindent\textit{Unbiased Teacher.}\quad   
    In the field of semi-supervised object detection (SS-OD), Unbiased Teacher (UT)~\cite{liu2022unbiased} addresses the bias issue in the pseudo-labeling, an inherent problem in the teacher-student learning framework without target domain labels. UT adopts the knowledge distillation approach as with AT and PT, while also balancing the loss to downweight overly confident pseudo-labels. 

    \noindent\textit{H$^{2}$FA R-CNN.}\quad
    In the presence of image-level annotations alone, termed Cross-domain Weakly Supervised Object Detection (CD-WSOD), H$^{2}$FA R-CNN~\cite{h2fa2022weakly} is a competitive work that aligns the backbone features in the image-level and the RPN and detection head in the instance-level remarkably well. The employed two-level coarse-to-fine alignment strategy provides an effective way to overcome the lack of fine annotations. We demonstrate in Sec.~\ref{sec:experiment} that this method is especially powerful in our targeted military settings without fine bounding box annotations.

    \begin{table*}[!t]
    \centering
    \caption{\textbf{Quantitative benchmark results on our proposed dataset pair} using the representative state-of-the-art domain adaptation strategies on our dataset. $^\dag$denotes COCO-pretrained weight initialization. Following the best-performing architecture per method for a fair comparison amongst each adaptation approach, $^{\ast}$means VGG backbone with random weight initialization, and for all other models, ImageNet pre-trained ResNet-101 backbone is used for initialization.}
    \label{tab:quantitative}
    \setlength\tabcolsep{2pt}
    \resizebox{.7\linewidth}{!}{
    \begin{tabular}{@{}rcccccc cca}
        \toprule
             & \multicolumn{6}{c}{Category (AP)}  && & \mc{1}{\;} \\
            \cmidrule{2-7}
            & Artillery & Drone & Helicopter & Soldier & Tank & Vehicle && mAP & mAP50 \\
        \toprule
            Source (Faster R-CNN~\cite{ren2015fasterrcnn}) & 0.3 & 8.2 & 3.1 & 3.3 & 2.9 & 16.9 && 5.8 & 15.3 \\ 
        \midrule
            \multicolumn{1}{l}{\textit{Unsupervised DA}} \\
            AT$^{\ast}$~\cite{li2022cross} & 0.0 & 21.7 & 2.1 & 0.0 & 0.0 & 1.7 && 4.3 & 10.6  \\
            PT$^{\ast}$~\cite{chen2022probabilisticteacher} & 0.2 & 4.6 & 14.8 & 0.2 & 3.4 & 4.1 && 4.5 & 13.8 \\ 
            
            CMT-AT$^{\ast}$~\cite{cao2023cmt} & 0.5 & 11.7 & 1.2 & 0.2 & 2.7 & 10.9 && 4.5 & 11.7 \\
            CMT-PT$^{\ast}$~\cite{cao2023cmt} & 0.0 & 6.1 & 17.2 & 0.0 & 2.0 & 1.9 && 4.6 & 11.3\\
        \midrule
            \multicolumn{1}{l}{\textit{Semi-supervised DA}} \\
            UT~\cite{liu2022unbiased} \tiny{(supervision$=$5\%)} & 0.6 & 10.6 & 2.4 & 4.5 & 10.6 & 21.3 && 8.3 & 21.5 \\
            UT~\cite{liu2022unbiased} \tiny{(supervision$=$100\%)} & 0.6 & 14.6 & 1.3 & 3.8 & 3.7 & 13.2 && 6.2 & 18.9 \\
        \midrule
            \multicolumn{1}{l}{\textit{Weakly supervised DA}} \\
            H$^{2}$FA R-CNN~\cite{h2fa2022weakly} & 46.6 & 41.2 & 52.4 & 16.7 & 47.3 & 38.0 && 40.4 & 73.3 \\
        \midrule
            \multicolumn{1}{l}{\textit{Supervised DA} \tiny{(Transfer Learning)}} \\
            EfficientDet$^\dag$~\cite{tan2020efficientdet} & 41.9 & 28.3 & 10.7 & 9.5 & 47.0 & 26.6 && 27.3 & 50.4 \\
            Faster R-CNN~\cite{ren2015fasterrcnn} & 41.3 & 47.5 & 80.5 & 29.1 & 54.6 & 56.0 && 51.5 & 80.1 \\
            Yolo V8x$^\dag$~\cite{ultralytics2023yolov8} & 51.4 & 52.7 & 79.4 & 28.1 & 60.6 & 65.0 && 56.2 & 84.5 \\
        \midrule
            Oracle (Faster R-CNN~\cite{ren2015fasterrcnn}) & 45.7 & 46.5 & 67.0 & 18.4 & 51.1 & 48.5 && 46.2 & 80.4 \\ 
        \bottomrule
        \end{tabular}
        }\\
    \end{table*}
\subsection{Datasets and Evaluation}

    Given the pair of synthetic and real dataset pair, we conducted our experiments to verify the performance of domain adaptation (Synthetic$\rightarrow$Real) methods on our datasets. 

    \noindent\textbf{Evaluation metric.} \quad
    We report the average precision (AP) of each of our defined military categories, as well as their mean score. Using the statistics from the confusion matrix, we follow previous work~\cite{li2022cross, chen2022probabilisticteacher, cao2023cmt, liu2022unbiased, h2fa2022weakly, ren2015fasterrcnn} to define the mean AP (mAP) as follows:
    
    \begin{equation}
        mAP = \frac{1}{C} \sum_{c=1}^{C} \textsc{AP}_{c},
    \end{equation}
    \noindent where \textsc{AP}$_{c}=\frac{1}{N} \sum_{n=1}^{N} \frac{TP}{TP+FP}\big\rvert_{n}$ denotes the class-wise average precision of all its samples, $N$, and C represents the number of classes (e.g., 6 in ours) in the dataset.

\subsection{Implementation Details}
    For a fair comparison with the previous methods, we use the standard Faster R-CNN object detector~\cite{ren2015fasterrcnn} with VGG-16~\cite{simonyan2014very} as the backbone for UDA methods and ResNet-101~\cite{he2016deep} as the backbone for the rest. For the UDA methods, we used a ResNet conv4 backbone with conv5 head as used in Faster R-CNN~\cite{ren2015fasterrcnn}. For the ResNet backbone, we use Deformable ConvNet~\cite{dai2017deformableconvnets}, which is a ResNet conv5 backbone with dilations in conv5, and standard conv and FC heads for mask and box prediction. We use Faster R-CNN since most algorithms have adopted this base framework (i.e., Detectron) and have demonstrated effectiveness in detection performance. We trained each model for at least 24K steps with a multistep linear warm-up scheduler, set the base learning rate to 2$\times10^{-3}$. We set the score threshold to 0.05 for evaluation and used ImageNet pre-trained weight initialization. For a fair comparative study across various adaptation methods, we evaluated all methods in the Detectron2~\cite{wu2019detectron2} framework and publicly available pre-trained weights using a single RTX A6000 GPU with a batch size of 4 on a PyTorch platform with CUDA 11.1. For all other hyperparameters, we follow those of the original work.

\subsection{Benchmark Result Analysis}    
    \begin{figure}[!t]
    \centering 
    \begin{subfigure}{0.47\linewidth}
    \centering
      \includegraphics[width=\linewidth]{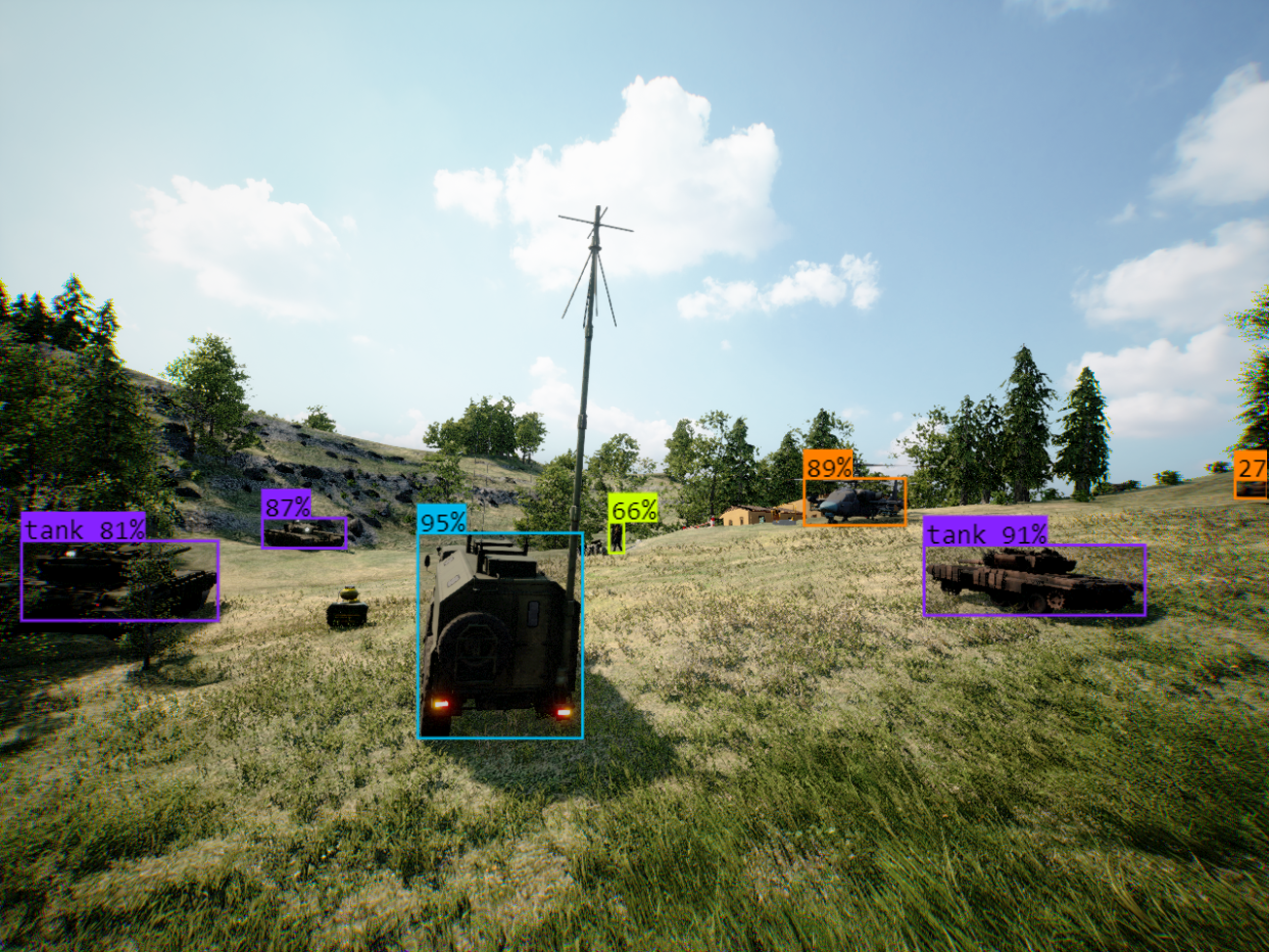}
        \caption{\textbf{Synthetic.} \\(purple: tank, blue: vehicle, orange: helicopter, green: soldier)}
        \label{fig:test_only_synthetic}
    \end{subfigure}\hfil 
    \begin{subfigure}{0.47\linewidth}
    \centering
      \includegraphics[width=\linewidth]{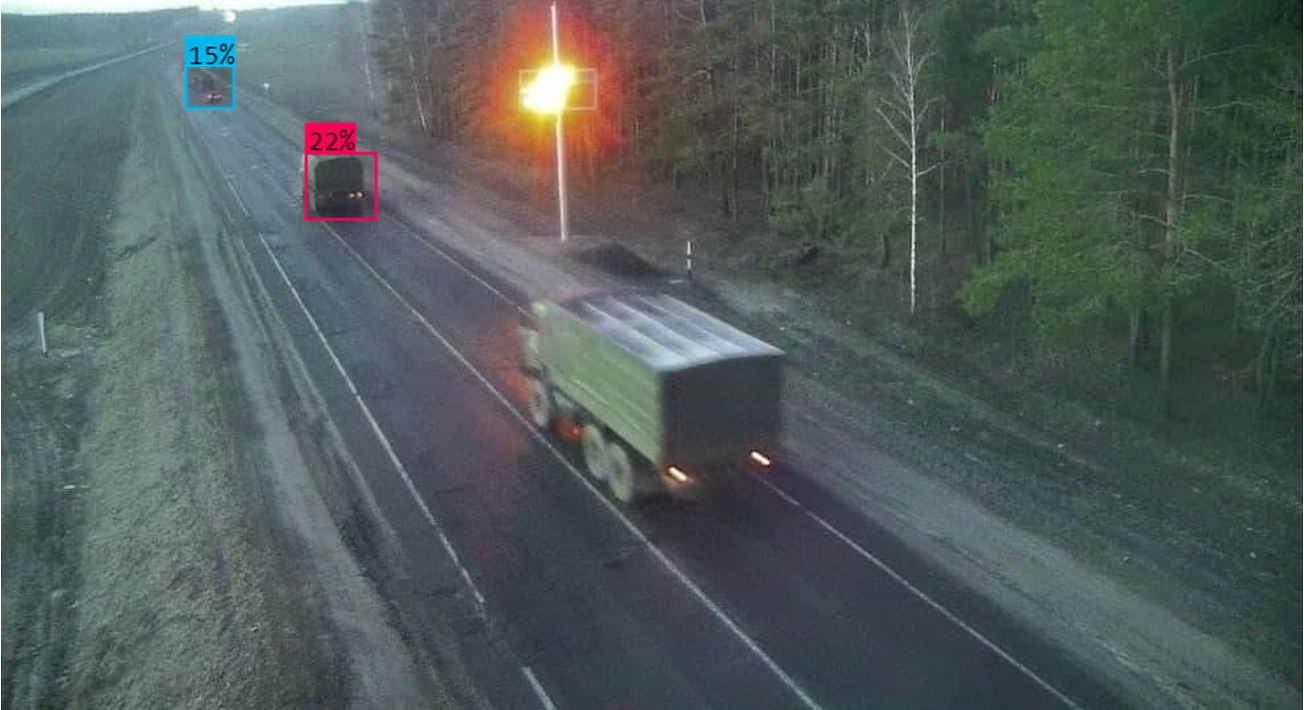}
        \caption{\textbf{Real.} \\(red: vehicle, blue: drone)\\}
        \label{fig:test_only_real}
    \end{subfigure}
    \caption{Single-source training results. Best in \textit{zoom}.}
    \label{fig:single_source_results}
    \end{figure}
    
    \begin{figure}[!t]
        \centering
        \includegraphics[width=\linewidth]{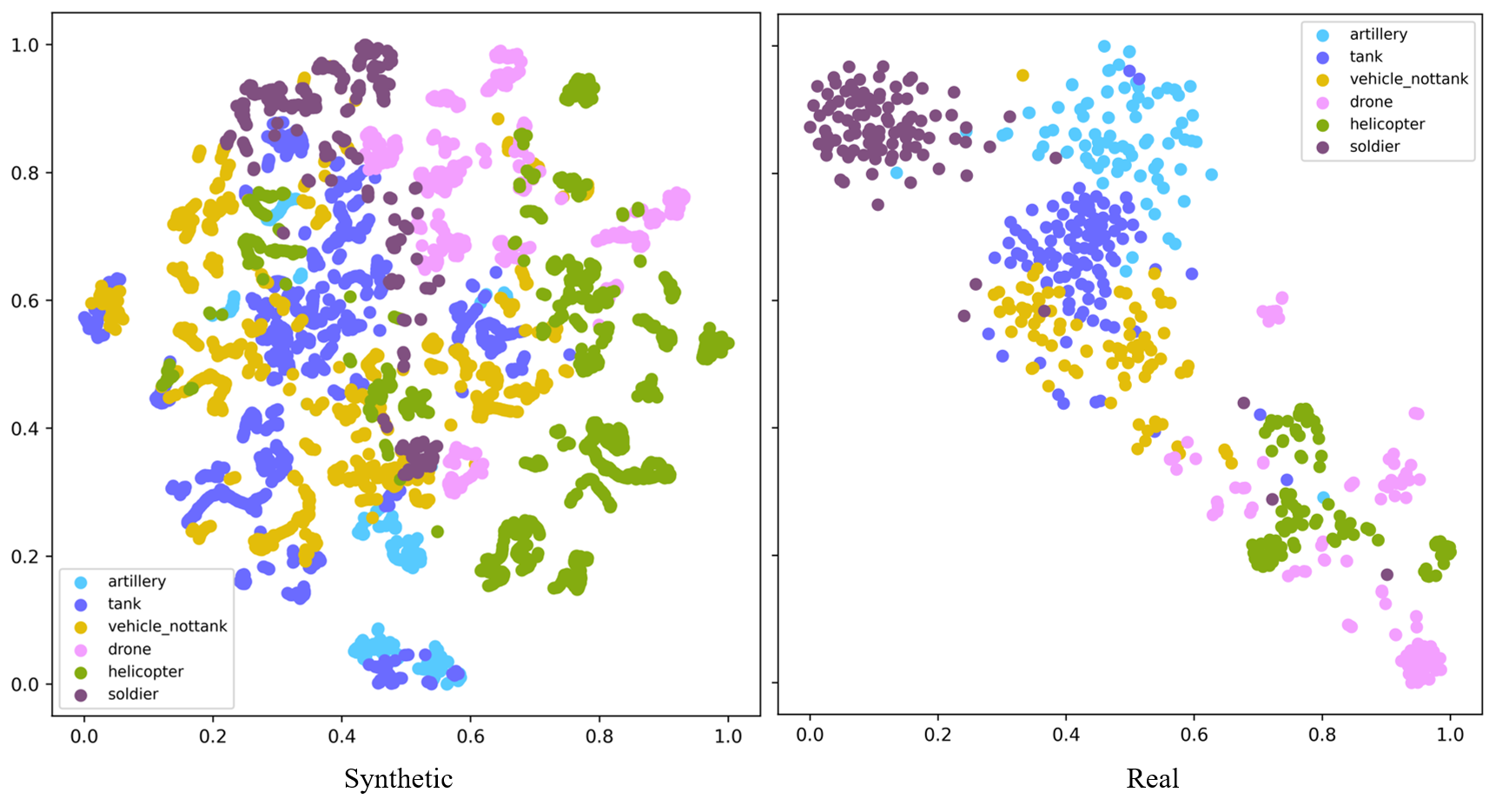}
        \caption{\textbf{t-SNE~\cite{van2008visualizingTSNE} visualization} of features extracted from ImageNet pre-trained ResNet-101. The class labels are clustered in different colors by the class label with the largest bounding box area in the image. \textit{Left}: our synthetic, and \textit{right}: real dataset pair.}
        \label{fig:tsne}
    \end{figure}
    
    \begin{figure}[!t]
        \centering
        \begin{subfigure}[b]{0.45\textwidth}
           \includegraphics[width=1\linewidth]{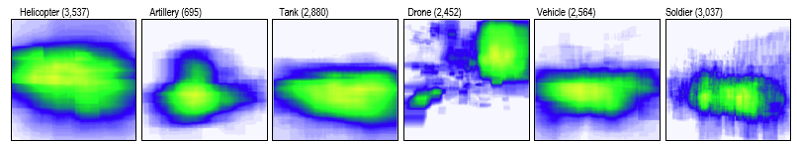}
           \caption{Synthetic Dataset}
           \label{fig:Ng1} 
        \end{subfigure}
        \begin{subfigure}[b]{0.45\textwidth}
           \includegraphics[width=1\linewidth]{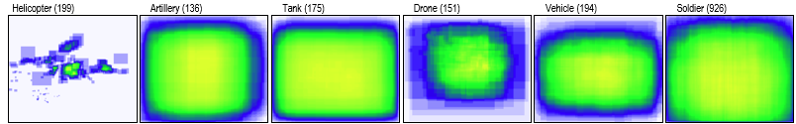}
           \caption{Real Dataset}
           \label{fig:Ng2}
        \end{subfigure}
        \caption{\textbf{Annotation heatmap distributions for the synthetic and real datasets}. Class-specific distributions are visualized for the six classes in the order of helicopter, artillery, tank, drone, vehicle and soldier (from \textit{left} to \textit{right}).}
        \label{fig:heatmap}
    \end{figure} 
    
    \begin{figure*}[!t]
        \centering
        \includegraphics[width=.85\paperwidth]{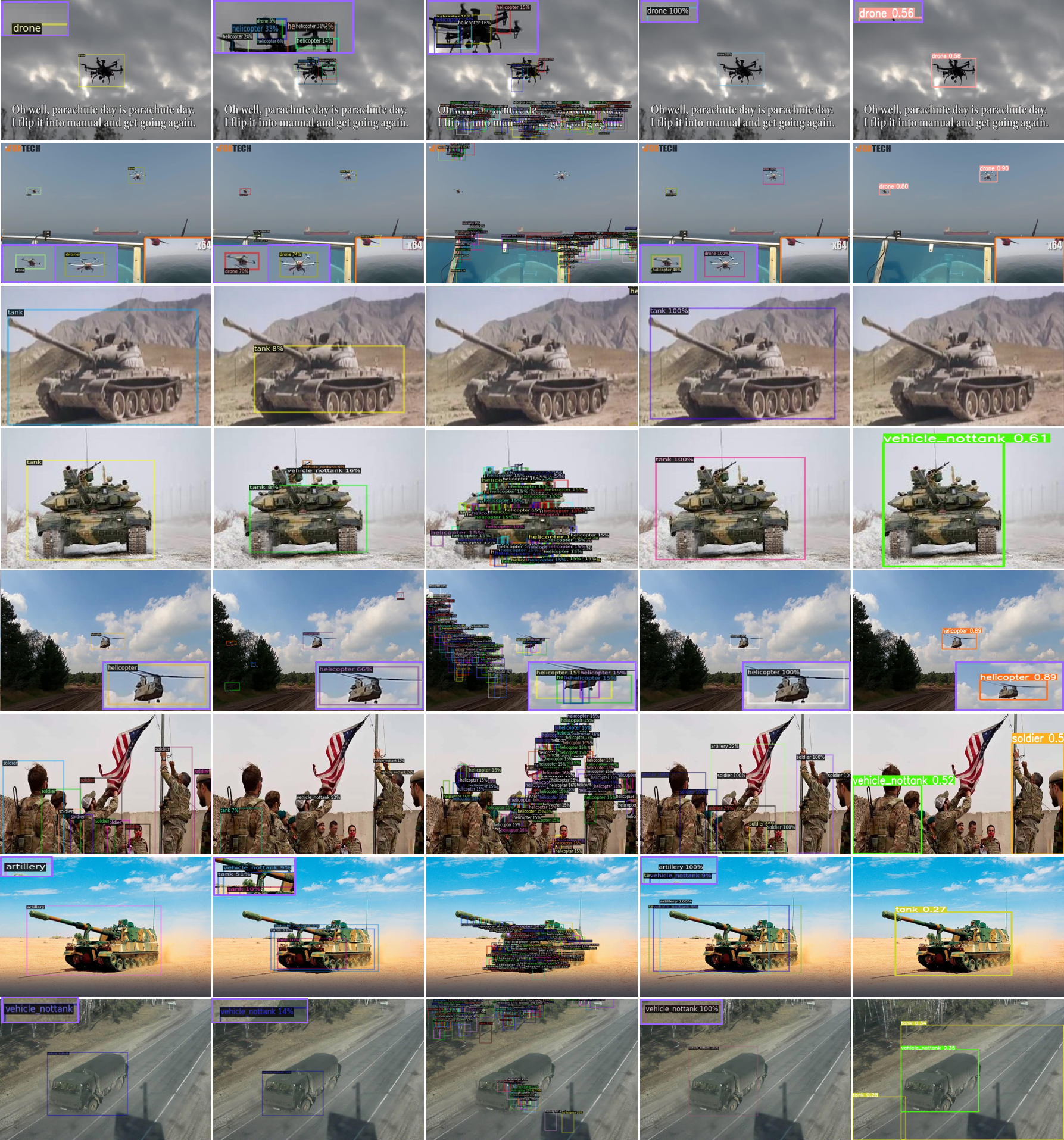}
        \caption{\textbf{Qualitative results.} We compare the detection results of unsupervised, semi-supervised, weakly supervised, and supervised domain adaptation methods evaluated on the real dataset ($\textsc{val}$). 
        Predictions are zoomed in \textcolor{violet}{purple} box. \textit{From left to right}: Ground Truth, CMT-PT (unsupervised DA), UT (semi-supervised DA), H$^{2}$FA R-CNN (weakly supervised DA), Yolo V8x (supervised DA). \textit{From top to bottom}: drone, drone, tank, tank, helicopter, soldier, artillery, vehicle. \textit{Best viewed in zoom and color}.}
        \label{fig:qualitative}
    \end{figure*}

    \noindent\textbf{Quantitative results.} \quad
    %
    %
    We report the experimental results for each class AP and mAP for each domain adaptation method in Table~\ref{tab:quantitative}. In the table, source and oracle each shows source and target domain training results on the target domain ({\sc{val}}). We observe that unsupervised adaptation methods significantly lack detection capability on the target domain after training with the labeled source and unlabeled target data together. For a deeper analysis of the detection failure, we resort to investigating the feature by t-SNE~\cite{van2008visualizingTSNE} visualization using an ImageNet pre-trained model in Fig.~\ref{fig:tsne} and annotations by heatmaps in Fig.~\ref{fig:heatmap}.
    
    In Fig.~\ref{fig:tsne}, we remark that each sample label shown in represents the class with the largest bounding-box area over all class objects present in each image. That is, the image is assumed to belong to a "tank" class if the tank object has the largest bounding-box area among other objects (e.g., helicopter or soldier). The same image labeling strategy was applied to the real set for a consistent comparison. Due to the nature of the detection task that involves multiple bounding-box labels in each image, we can only observe and relatively compare the general feature distributions specific to each dataset as shown in Fig.~\ref{fig:tsne}. Therefore, the figure serves to provide insight into how each training (synthetic) and validation (real) set lies in the feature space based on the class object with the largest area. In that context, the t-SNE plot confirms that the synthetic set is as challenging as the real, since some of the class sample features overlap more than the real samples. We designed the synthetic set as such to ensure good performance on the target real set. 
    
    We noticed that the class-wise synthetic data features are more spread and intermingled with each other, while the real data features are class-wise clustered. Given these two vastly different data distributions, the network struggles to meet the unsupervised domain adaptation objective of high detection mAP. Although the ResNet-extracted features appear to be rather more difficult to distinguish for the synthetic and the real, upon training on single-source (synthetic-only and real-only) scenarios, we verify with the promising single-source training performance scores that the features in Fig.~\ref{fig:tsne} necessarily harm domain adaptive training strategies. We present the single-source training results in Fig.~\ref{fig:single_source_results}, with the quantitative results as follows: $\textrm{mAP=}99.2\%, \textrm{precision=}97.5\%, \textrm{recall=}97.6\%$ for synthetic-only, and $\textrm{mAP=}88.8\%, \textrm{precision=}88.1\%, \textrm{recall=}83.7\%$ for real-only.

    Furthermore, Fig.~\ref{fig:heatmap} exhibits real annotation distributions that are far too spread out by class. The annotation heatmaps for the synthetic dataset are intuitive from natural visual shapes by category, but those for the real data are indistinguishable from each other because of too diverse object positions, hinting at the difficulty of detecting real sample objects with training on our synthetic data.    
    
    \noindent\textbf{Qualitative results.} \quad
    We also provide detection visualizations to intuitively demonstrate the benefit of each representative method in Fig.~\ref{fig:qualitative}.
    We compare unsupervised, semi-supervised, weakly supervised, and supervised DA results side-by-side to effectively highlight the differences in detection quality.
    We find that unsupervised DA (CMT-PT~\cite{cao2023cmt})) is able to minimally identify and distinguish foreground objects, but the confidence score is relatively low and flying objects are often overly misclassified as helicopters. On the other hand, the semi-supervised method (UT~\cite{liu2022unbiased}) has a high rate of false positive, as its detection boxes are not uncertain in most images, thus it is difficult to determine which object it is classified as. Weakly supervised method (H$^{2}$FA R-CNN), on the contrary to the two previous ones, detects the object boundaries well with slight difficulty in capturing the correct bounding box entirely. Yolo V8x, after fine-tuning, detects as well as H$^{2}$FA R-CNN, but with lower confidence scores and a few failure cases, such as on the tank class object.

    \noindent\textbf{Limitations.} \quad 
    From Table~\ref{tab:quantitative}, there are certain classes (\textit{e.g.,} \textit{Artillery} and \textit{Soldier}) that are difficult to be predicted well. We attribute these challenging classes to the lack of adequate military target assets and our design aim of building a fair and even dataset during the construction process.
    For instance, \textit{Soldier} class, along with the other classes, contains around ~3\textit{k} annotated instances, except for \textit{Artillery} class for which we were not able to find as many distinctive and adequate Unreal Engine assets. Furthermore, the \textit{Soldier} class in the real set has the largest sample size among all six classes of interest. We attribute the poor performance for the \textit{Soldier} class, in particular, to two factors: (1) small object size, and (2) blended instances with the background (e.g., trees). As aforementioned in Sec.~\ref{sec:dataset}, we removed data samples that do not meet the size, redundancy, recognizability, or valid class criteria. For the \textit{Soldier} class samples, the we mostly removed redundant samples to prevent overfitting to certain scenes and to keep from possible bias toward soldier instances for fair dataset instances as described in Table 2 in the manuscript. While maintaining fairness in synthetic, we had the largest number of Soldier samples in the real dataset to accommodate the difficulty due to size \& background blending, particularly present in the Soldier class.



\section{Conclusion}
\label{sec:conclusion}

In this paper, we have presented a novel synthetic dataset for armored military target detection, as well as a paired web-collected real dataset for evaluation. While a number of military datasets have been presented, there is still a lack of publicly open armored detection datasets for cost-effective RGB sensor-based systems. As real-world battlefield scenes require real-time, accessible, and accurate detection systems, satellite data-based systems cannot always accompany target detection with high enough accuracy and precision. Moreover, with varying environmental domains, no other work has delved into cross-domain adaptation methods for military target detection. Hence, in this paper, we explore the potential of using synthetic data\footnote{dataset will be openly available later upon acceptance.} as training data for use in real domains. Although current unsupervised domain adaptation methods demonstrate that UDA approach is still inadequate in performance, we found that adding even weak supervision significantly boosts the detection quality through comparative studies. Although the datasets compared in domain adaptation methods resemble each other to a moderate degree, military-specific environments vary significantly. Therefore, we conjecture that a domain generalizable object detection method may be necessary for military target detection in the future. 




\bibliographystyle{IEEEtran}
\bibliography{references}

\begin{IEEEbiography}[{\includegraphics[width=1in,height=1.25in,clip,keepaspectratio]{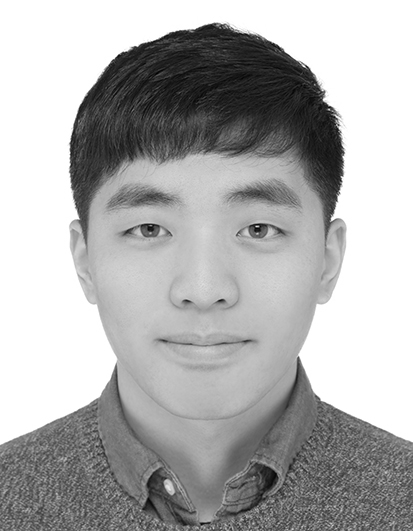}}]{Jongoh Jeong} (Graduate Student Member, IEEE) received the B.E. degree in electrical engineering from The Cooper Union for the Advancement of Science and Art, New York, NY, in 2020 and the M.S. in electrical engineering at Korea Advanced Institute of Science and Technology (KAIST), Daejeon, Republic of Korea, in 2022. He is currently pursuing a Ph.D. degree in the Robotics Program at KAIST. His current research interests include domain adaptation and generalization, metalearning, optimization methods, and continual learning related to vision-based autonomous driving.
\end{IEEEbiography}
\begin{IEEEbiography}[{\includegraphics[width=1in,height=1.25in,clip,keepaspectratio]{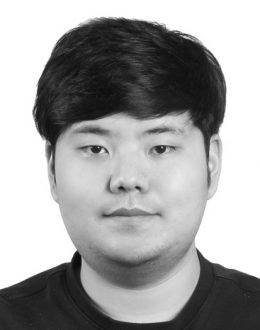}}]{Youngjin Oh} received the B.S. degree in Electrical and Information Engineering from Seoul National University of Science and Technology in 2023. He is currently pursuing the M.S. degree in the Division of Future Vehicle at KAIST. His research interests include occupancy prediction and autonomous driving.
\end{IEEEbiography}
\begin{IEEEbiography}[{\includegraphics[width=1in,height=1.25in,clip,keepaspectratio]{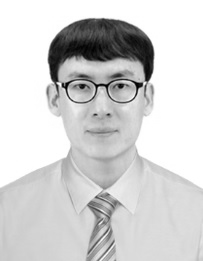}}]{GYEONGRAE NAM} received the B.S. degree in Computer Science from the Kyung Hee University, Seoul, South Korea, in 2007 and the M.S. degree in Software Security from Korea University, Seoul, South Korea in 2020. Since 2007, he has been a Senior Researcher with LIG Nex1, Seongnam, South Korea. His research interests include multi-domain control of unmanned aerial vehicle, Dynamic Mission Planning and Architecture/Interface/Framework of Unmanned GCS. 
\end{IEEEbiography}
\begin{IEEEbiography}[{\includegraphics[width=1in,height=1.25in,clip,keepaspectratio]{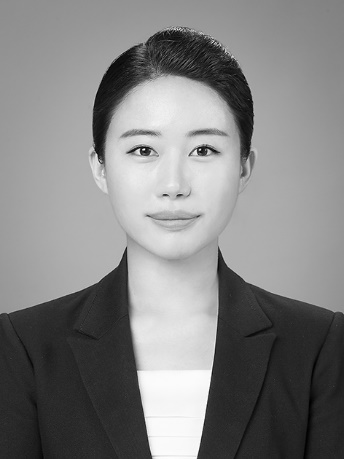}}]{JEONGEUN LEE} received the master’s degree in Computer Science from Hanyang University, Seoul, South Korea, in 2021. Since 2021, she has been a Senior Researcher with LIG Nex1, Seongnam, South Korea. Her research interests include the control of unmanned aerial vehicle and object detection.
\end{IEEEbiography}
\begin{IEEEbiography}[{\includegraphics[width=1in,height=1.25in,clip,keepaspectratio]{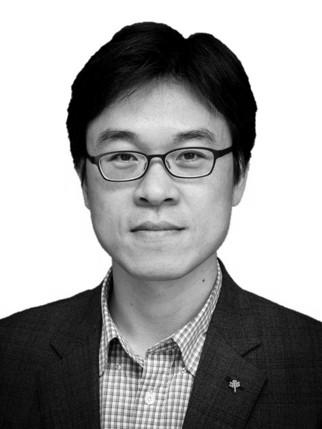}}]{Kuk-Jin Yoon} received the B.S., M.S. and Ph.D. degrees in electrical engineering and computer science from the Korea Advanced Institute of Science and Technology in 1998, 2000, and 2006, respectively. He is now an Associate Professor at the Department of Mechanical Engineering, Korea Advanced Institute of Science and Technology (KAIST), South Korea, leading the Visual Intelligence Laboratory. Before joining KAIST, he was a Postdoctoral Fellow in the
PERCEPTION Team, INRIA, Grenoble, France, from 2006 to 2008, and was an Assistant/Associate Professor at the School of Electrical Engineering and Computer Science, Gwangju Institute of Science and Technology, South Korea, from 2008 to 2018. His research interests include various topics in computer vision such as multi-view stereo, visual object tracking, SLAM and structure-from motion, 360 camera and event-camera-based vision, and sensor fusion.
\end{IEEEbiography}

\EOD

\end{document}